\documentclass{article}
\usepackage{amsmath}
\usepackage{graphicx}
\usepackage{booktabs}
\usepackage{multirow}
\usepackage{multicol}
\usepackage{cite}
\usepackage{geometry}
\usepackage{indentfirst}
\usepackage{bm}
\title{MLPSVM:A new parallel support vector machine to multi-label learning}
\author{\bf{Yanghong Liu,Jia Lu,Tingting Li}\\\fontsize{8pt}{5em}\selectfont CHONGQING NORMAL UNIVERSITY\\liuyanghong@cqnu.edu.cn}
\begin{document}\large
\maketitle
\begin{abstract}\large
Multi-label learning has attracted the attention of the machine learning community. 
The problem conversion method Binary Relevance converts a familiar single label into a multi-label algorithm.
The binary relevance method is widely used because of its simple structure and efficient algorithm. 
But binary relevance does not consider the links between labels, making it cumbersome to handle some tasks. 
This paper proposes a multi-label learning algorithm that can also be used for single-label classification.
It is based on standard support vector machines and changes the original single decision hyperplane into two parallel decision hyperplanes,
which call multi-label parallel support vector machine(MLPSVM).At the end of the article, MLPSVM is compared with other multi-label learning algorithms. 
The experimental results show that the algorithm performs well on data sets.
    
\end{abstract}
\section{Introduction}
Multi-label learning is widely used to serve our real life\cite{Montanari2016Selectivity,Mercan2017Multi,Wan2016Mem,Guo2016Human,Wan2017Transductive,Zhang2018Ontological}. 
In the field of music classification\cite{Oramas2017Multi}, traditional classification algorithms will classify music into pop and rock classes,
which are too broad. Use multi-label learning algorithms to expand music classification to tasks with multiple labels will make music classification more in line with practical requirements. 
For protein chloroplast localization\cite{Wan2017Transductive} in the biological field, the traditional method uses single-position protein chloroplast localization, but ignores multi-position protein chloroplast localization.
Using multi-label learning to perform multi-position protein chloroplast localization has better effect. Studying multi-label learning algorithm will help to solve the existing problems.
Zhang\cite{zhang2013review},has reviewed multi-label learning and introduced multi-label learning in detail.
Multi-label learning algorithms are simply divided into the following two categories:
\begin{itemize}
\item Problem conversion methods:converts the multi-label learning problem into a familiar single label problem to solve the problem, such as the first-order method Binary Relevance and the high-order method Classiﬁer Chains.
\item Algorithm adaptive method: this kind of algorithm directly uses multi-label data to solve the multi-label problem. The representative algorithms are the first-order ML-KNN\cite{MLKNN}, ML-DT\cite{MLDT} second-order Rank-SVM\cite{ranksvm}, CML\cite{cml}. 
\end{itemize}\par
In multi-label learning, considering the association between labels, there will be label combination explosion\cite{Zhang2010Multi} and redundant feature information\cite{Xu2016Multi}. 
Many multi-label learning algorithms are proposed\cite{Read2017Multi,Zhu2017Multi,Babbar2017DiSMEC,TRAJDOS201860,7869900,Meng2017An} to solve the above problems. 
For multi-label data, the data has both the same label and different labels, which will result in data crossing.
It seems that few scholars pay attention to the situation of data crossing. Chen\cite{MLTSVM} proposed MLTSVM,
which can handle crossing amount data well.\par
In this paper, a multi-label classification method based on support vector machine(SVM)\cite{SVM} is designed call it MLPSVM.
MLPSVM uses two parallel hyperplanes to identify a label.
The use of MLPSVM will facilitate SVM to perform multi-label classification tasks.
At the same time, MLPSVM is a convex quadratic programming model.
\subsection{Notation and Setup}
Let $X=R^{n}$ be the n-dimensional input space and $Y=R^{d}$ be the d-dimensional label space,where $y=(y_{1},y_{2},y_{3}...y_{d})\in\{1,-1\}^{d}$.
Given a multi-label training set $D=\{(x_{i},y_{i})|1\leq i \leq m\}$,where $x_{i}\in X$ is a feature vector and $y_{i}\in Y$ is the set of label associateal with the $x_{i}$.
The goal of MLPSVM is to obtain two matrices
    \begin{equation*}
        w=\begin{bmatrix}
            w_{11} & w_{12}& w_{13} & \dots & w_{1n} \\
            \hdotsfor{5} \\
            w_{j1} & w_{j2} & w_{j3} & \dots & w_{jn} \\
            \hdotsfor{5}\\
            w_{d1} & w_{d2} & w_{d3} & \dots & w_{dn}
          \end{bmatrix}
        ,b=\begin{bmatrix}
            b_{11}&{b_{12}}\\
            \hdotsfor{2}\\
            b_{j1}&b_{j2}\\
            \hdotsfor{2}\\
            b_{d1}&b_{d2}
          \end{bmatrix},
      \end{equation*}
$w_{j}$ is row j matrix $w$.For the input feature vector $x\in X$.
The label $y=(y_{1},y_{2},y_{3}...y_{d})\in Y$ about $x$  is  obtained by expression 
\begin{flalign}    
y_{j}=\Bigg \{_{-1\ otherwise}^{+1\ f_{j1}=w_{j}x+b_{j1}\geq 0 \ and\  f_{j2}=w_{j}x+b_{j2}\leq 0}.
\end{flalign}
\section{Support Vector Machine}
\begin{figure}[htb]
    \centering
    \includegraphics[height=200pt,width=300pt]{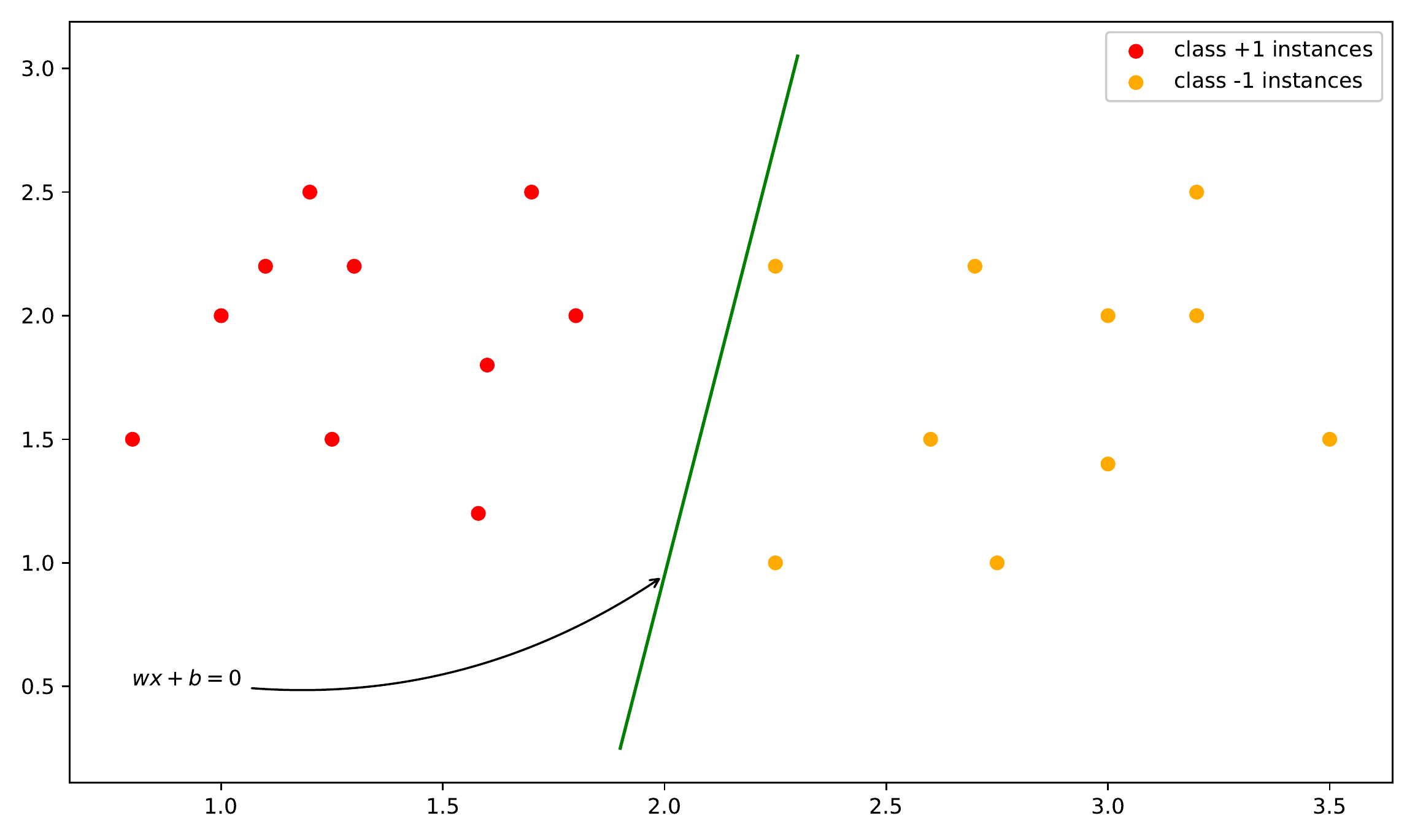}
    \caption{An intuitive geometric interpretation SVM}
    \label{SVM}
\end{figure}

Support vector machine is a binary classification model. Its basic model is the linear classifier with the largest interval defined on the feature space.
The learning strategy of support vector is to find a hyperplane in the sample space to separate the two classes at maximum intervals.
We denote the set of training data as $ T=\{(x_{i},y_{i})|1\leq i \leq m\}$ where $x_{i}\in R^{n}$
represents an input instance with the corresponding label $y_{i}\in \{1,-1\}$.The primal problem of SVM can be expressed as 
\begin{flalign}
    \mathop{min}\limits_{w,b,\delta_{i}}\quad  &\frac{1}{2}\Vert w \Vert^{2}+c\sum\limits_{i=1}^{m}\delta_{i} \nonumber\\ 
    s.t.\quad&y_{i}(wx_{i}+b)\geq 1-\delta{i}, \nonumber \\
        &\delta{i}\geq 0,i=1,2,...,m, \label{svm}
\end{flalign}
where $\delta_{i}$ is the slack variable to indicate the misclassification error,and $c>0$ is the penalty parameter.
$w$ and $b$ are the normal vector and the bias term of hyperplanes respectively.
An intuitive geometric interpretation for SVM is shown in \ref{SVM}.Formula \ref{svm} is a convex quadratic programming problem,
$w$ and $b$ can be obtained by solving this problem.Therefore,the desision function $f(x)=sign(wx+b)$.
\par
A new data instance $x$ ,the class $y={+1,-1}$ about x is obtained by expressed
\begin{flalign}    
    y=\Bigg\{_{+1 \   wx+b\geq0.}^{-1 \     wx+b<0}
\end{flalign}
Introducing kernel techniques into support vector machines can also make support vector machines nonlinear classifiers.
\section{Multi label parallel support vector machine}
\begin{figure}[htb]
    \centering
    \includegraphics[height=200pt,width=300pt]{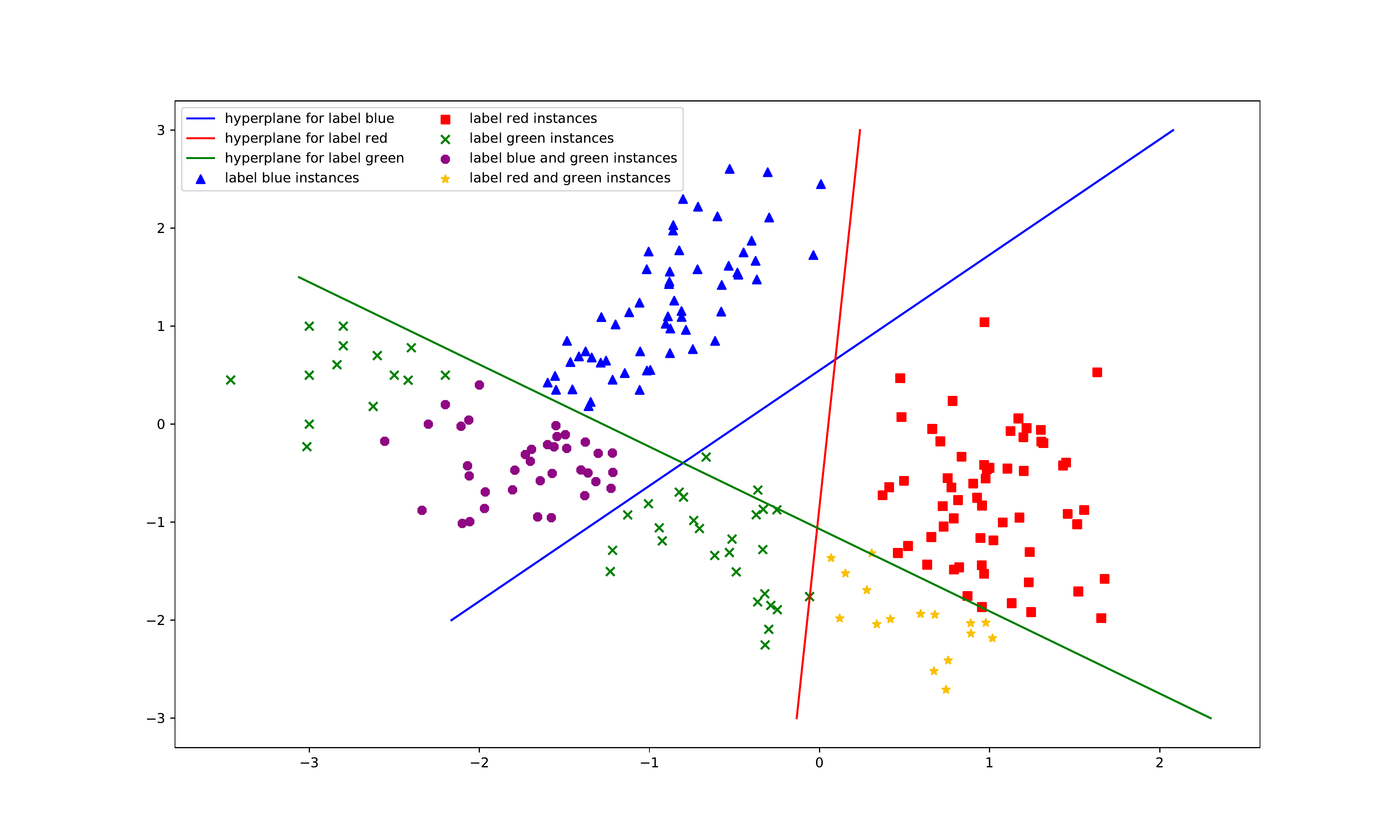}
    \caption{An intuitive geometric interpretation BR\_{SVM}}
    \label{MLSVM}
\end{figure}

This section introduces multi-label parallel support vector machine.
Considering the data intersection in the multi-label problem, a multi-label parallel support vector machine is designed to solve the problem.
As shown in Figure \ref{MLSVM}, using BR\_SVM will cause the data with only green labels on the left to be incorrectly predicted with blue labels.
\par
MLPSVM uses two parallel hyperplanes $f_{j1}(x)=w_{j}x+b_{j1}$ and $f_{j2}(x)=w_{j}x+b_{j2}$ to identify a tag. 
The goal is to locate the relevant tag data between the two hyperplanes.
The function of hyperplane $f_{j1}$ is still the same as that of traditional support vector machine, which is responsible for separating different label data, 
so that for all sample data there are
\begin{flalign}    
    y_{ij}(w_{j}x+b_{j1})\geq 1-\delta_{i}.
\end{flalign}
The hyperplane $f_{j2}$ is responsible for locating the corresponding data between the two hyperplanes $f_{j1}$ and $f_{j2}$. 
For data x, if $y_{ij}=+1$, there are $f_{j1}(x)\geq0$, $f_{j2}(x)\leq0$. Design the following constraints
\begin{flalign}    
    (1+y_{ij})(w_{j}x+b_{j1})\leq 0. 
    \label{eq}
\end{flalign}
Constraint \eqref{eq} makes the corresponding data with positive label $y_{ij}=+1$ subject to the above constraint.
For data with negative label $y_{ij}=-1$, because $(1+y_{ij})(w_{j}x+b_{j1})\leq 0$, this is an identity and will not be constrained.
In order to enable hyperplanes $f_{j1}$ and $f_{j2}$ to tightly surround data, a relaxation variable is added in constraint \ref{eq},
so that constraint \eqref{eq} becomes an equality constraint
\begin{flalign}    
    (1+y_{ij})(w_{j}x+b_{j1})+\alpha_{ij}=0. 
    \label{eq1}
\end{flalign}
The $\alpha_{ij}$ are slack variables measureing the  distance that hyperplanes $f_{j2}$ makes on the data.
Bring $\alpha_{ij}$ into the objective function, there are
\begin{flalign}    
    \mathop{min}\limits_{w,b,\delta,\alpha}\frac{1}{2} \sum\limits_{j=1}^{d}\Vert{w_{j}}\Vert^{2}+C_{1}\sum\limits_{i=1}^{m}\sum\limits_{j=1}^{d}\delta_{ij}+C_{2}\sum\limits_{i=1}^{m}\sum\limits_{j=1}^{d}\alpha_{ij} 
    \label{eq2}
\end{flalign}
\begin{figure}[h]
    \centering
    \includegraphics[height=200pt,width=300pt]{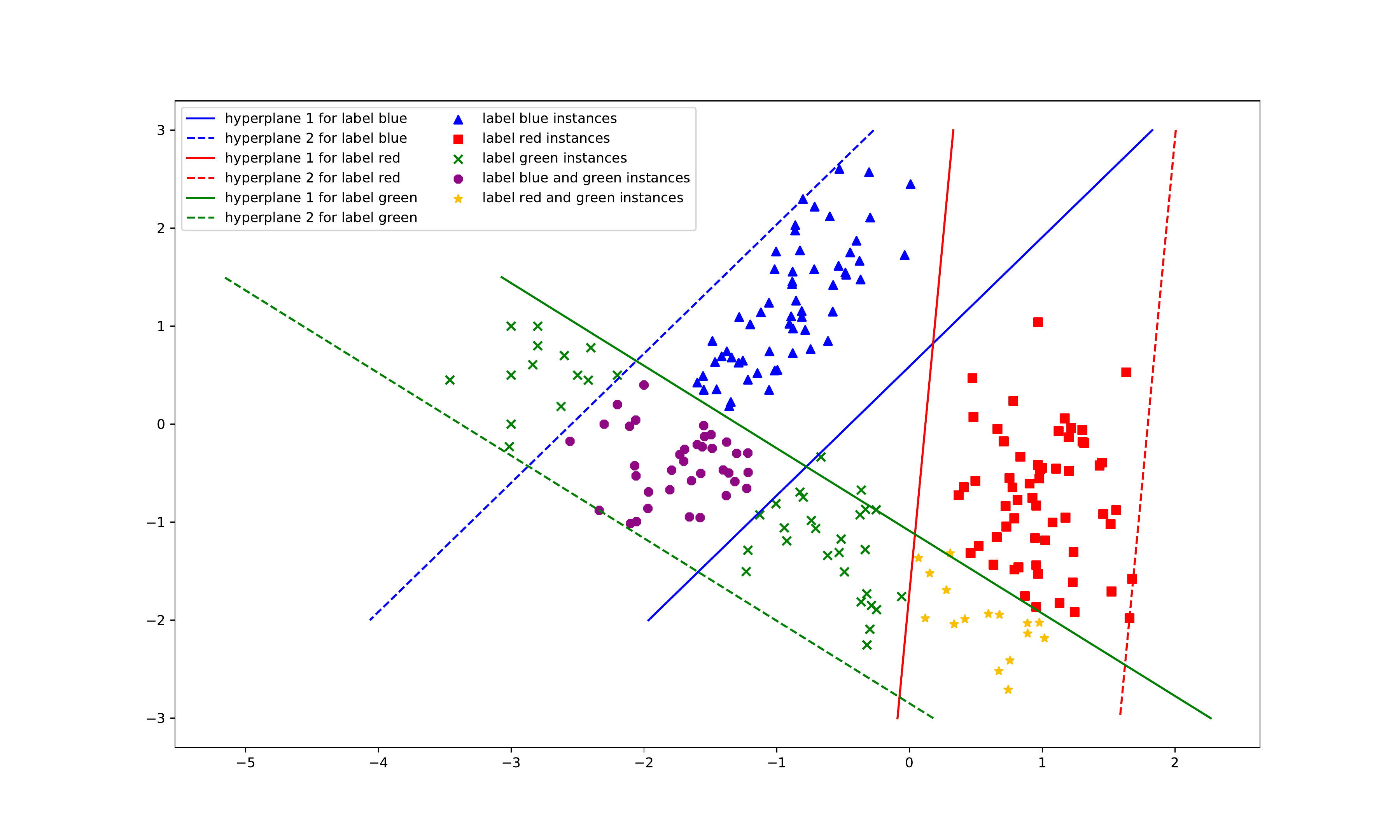}
    \caption{An intuitive geometric interpretation MLPSVM}
    \label{MLPSVM}
\end{figure}\par
In this way, MLPSVM will consider the distance of data to hyperplane $f_{j2}$.An intuitive geo-metric interpretation for MLPSVM is shown in Figure \ref{MLPSVM}.
\subsection{linear multi label parallel support vector machine}
Now begin introduce  parallel multi label support vector machine classifier by formulating the classification problem as:\\
\begin{flalign}
\mathop{min}\limits_{w,b,\delta,\alpha}\quad \frac{1}{2} &\sum\limits_{j=1}^{d}\Vert{w_{j}}\Vert^{2}+C_{1}\sum\limits_{i=1}^{m}\sum\limits_{j=1}^{d}\delta_{ij}+C_{2}\sum\limits_{i=1}^{m}\sum\limits_{j=1}^{d}\alpha_{ij} \nonumber \\
s.t.\quad&y_{ij}(w_{j}x_{i}+b_{j1})\geq1-\delta_{ij} \nonumber \\
&(1+y_{ij})(w_{j}x_{i}+b_{j2})+\alpha_{ij}=0 \nonumber \\
&\delta_{ij}\geq0,\alpha_{ij}\geq0 \nonumber \\
&for\ i=1,...,m \ and\ j=1,...,d  . 
\label{eq3}
\end{flalign}
\begin{figure}[htb]
    \centering
    \includegraphics[height=250pt,width=350pt]{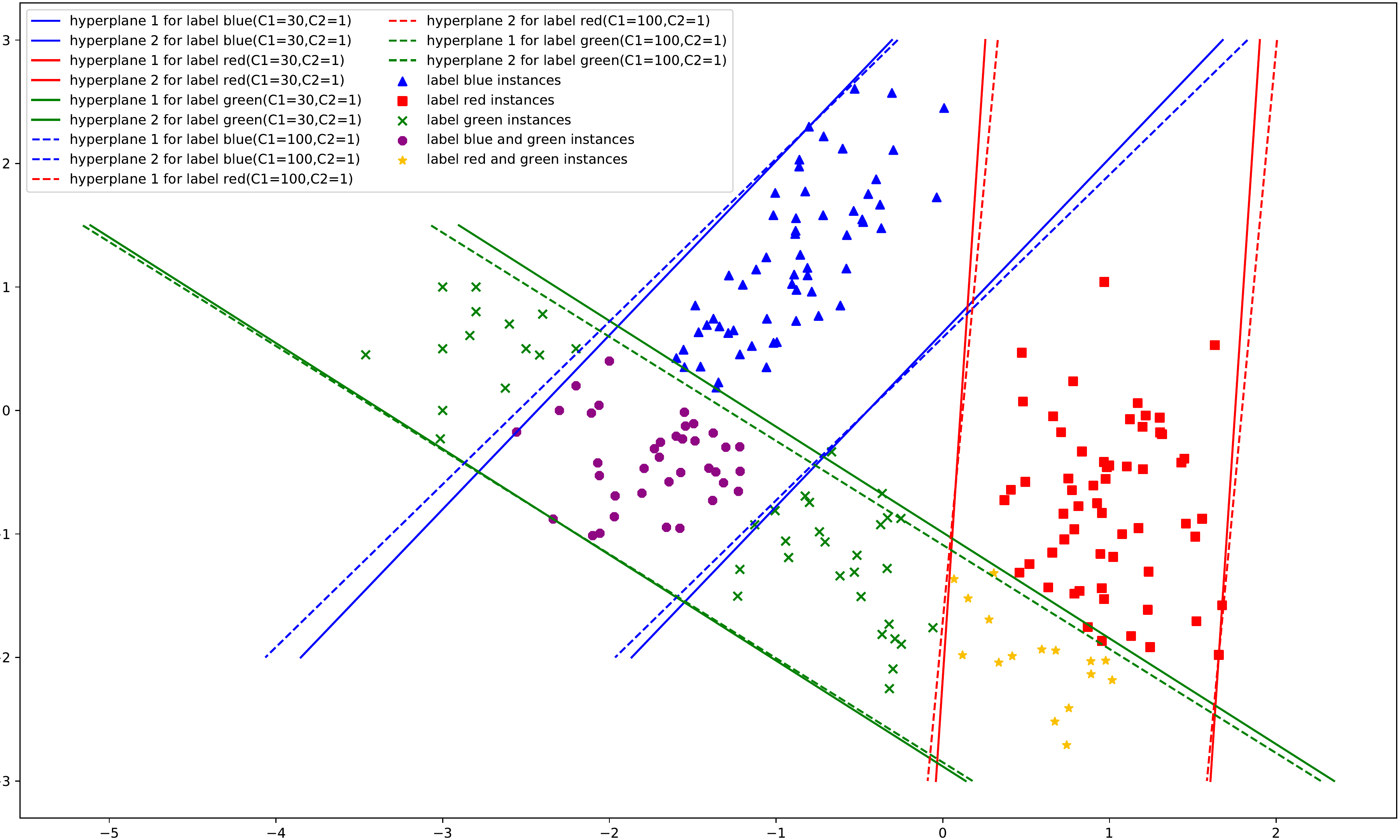}
    \caption{MLPSVM use two ratios $\frac{C_{1}}{C_{2}}$, for the same training data.}
    \label{figure2}
\end{figure}\par
In this problem, $C_{1}$ and $C_{2}$ are regularization parameters.
The $\delta_{ij}$ are slack variables measureing the error that hyperplanes $f_{j1}$ makes on the data.
The $\alpha_{ij}$ are slack variables measureing the  distance that hyperplanes $f_{j2}$ makes on the data.
This constraint ? can only be applied to positive labels.
For negative labels $y_{ij}=-1$,$(1+y_{ij})(w_{j}x_{i}+b_{j2})+\alpha_{ij}=0$ which is an identity.
The emphasis of the model is adjusted by changing the value of the ratio $\frac{C_{1}}{C_{2}}$.
The larger the ratio $\frac{C_{1}}{C_{2}}$, the better the hyperplane $f_{j1}$ can separate the positive training data from the negative training data.
The smaller the ratio $\frac{C_{1}}{C_{2}}$, the closer the positive training data will be to the hyperplane $f_{j2}$.
As show in Figure \ref{figure2},use two ratios $\frac{C_{1}}{C_{2}}$, for the same training data.\par
According to the above model, observation shows that the model can be simplified,$\Vert{w_{j}}\Vert^{2}$ and $w_{j}x_{i}$\ Appears in the model.$w_{j}$\ is row j of matrix w.
this will bring difficulties to the solution.Through the following expression
\begin{flalign}
    W=(w_{1},w_{2},w_{3},...,w_{d}) \\
    \Phi(x_{i},j)=(\underbrace{0,...,0}_{j-1},x,\underbrace{0,...,0}_{d-j})
\end{flalign}
where we have denoted by 0 the vector in $R^{d}$\ whose coordinates are all zero.By construction we have that\\
\begin{flalign}
    \frac{1}{2}\sum\limits_{j=1}^{d}\Vert{w_{j}}\Vert^{2}=\frac{1}{2}\Vert{W}\Vert^{2} \label{eq40} \\
     w_{j}x_{i}=W\cdot\Phi(x_{i},j). \label{eq41}
\end{flalign}
Combining equations \eqref{eq40},\eqref{eq41}and \eqref{eq3} respectively will yield the following formula\\
\begin{flalign}
    \mathop{min}\limits_{W,b,\delta,\alpha}\quad &\frac{1}{2}\Vert{W}\Vert^{2}+C_{1}\sum\limits_{i=1}^{m}\sum\limits_{j=1}^{d}\delta_{ij}+C_{2}\sum\limits_{i=1}^{m}\sum\limits_{j=1}^{d}\alpha_{ij} \nonumber \\ 
    s.t.\quad &y_{ij}(W\cdot \Phi(x_{i},j)+b_{j1})\geq1-\delta_{ij}  \nonumber\\
    &(1+y_{ij})(W\cdot \Phi(x_{i},j)+b_{j2})+\alpha_{ij}=0 \nonumber\\
    &\delta_{ij}\geq0,\alpha_{ij}\geq0 \nonumber \\
    &for\ i=1,...,m \ and\ j=1,...,d   \label{eq4}
\end{flalign}
The new expression obtained is similar to the standard SVM expression.
At the same time, it can be easily seen that this is a convex quadratic programming problem. 
\subsection{Dual Optimization Problem}
An importtant characteristic of SVM is that they can be used to estimate highly non-linear functions through the use of kernels.
In the section,derive the dual of problem \eqref{eq4}.The Lagrangian with $\eta_{ij}\geq 0 ,\lambda_{ij}\geq 0,\mu_{ij} \geq 0$\ of (\ref{eq4}) is
\begin{flalign}
L(W,b,\delta,\alpha,\eta,\theta,\lambda,\mu)=&\frac{1}{2}\Vert{W}\Vert^{2}+C_{1}\sum\limits_{i=1}^{m}\sum\limits_{j=1}^{d}\delta_{ij}+C_{2}\sum\limits_{i=1}^{m}\sum\limits_{j=1}^{d}\alpha_{ij} \nonumber \\
&-\sum\limits_{i=1}^{m}\sum\limits_{j=1}^{d}\eta_{ij}(y_{ij}(W\cdot \Phi(x_{i},j)+b_{j1})-1+\delta_{ij}) \nonumber \\
&+\sum\limits_{i=1}^{m}\sum\limits_{j=1}^{d}\theta_{ij}((1+y_{ij})(W\cdot \Phi(x_{i},j)+b_{j2})+\alpha_{ij}) \nonumber \\
&-\sum\limits_{i=1}^{m}\sum\limits_{j=1}^{d}\lambda_{ij}\delta_{ij}-\sum\limits_{i=0}^{m}\sum\limits_{j=0}^{d}\mu_{ij}\alpha_{ij}.
\end{flalign}
Then according to
\begin{flalign}
    &\frac{\partial L}{\partial W}=W-\sum\limits_{i=1}^{m}\sum\limits_{j=1}^{d}\eta_{ij}y_{ij}\Phi(x_{i},j) + \sum\limits_{i=1}^{m}\sum\limits_{j=1}^{d}\theta_{ij}(1+y_{ij})\Phi(x_{i},j)=0 \\
    &\frac{\partial L}{\partial b_{j1}}=-\sum\limits_{i=1}^{m}\eta_{ij}y_{ij}=0,\forall j=1,...,d\\
    &\frac{\partial L}{\partial b_{j2}}=-\sum\limits_{i=1}^{m}\theta_{ij}(1+y_{ij})=0,\forall j=1,...,d\\
    &\frac{\partial L}{\partial \delta_{ij}}=C_{1}-\eta_{ij}-\lambda_{ij}=0,\forall i=1,...,m,\forall j=1,...,d \\
    &\frac{\partial L}{\partial \alpha_{ij}}=C_{2}+\theta_{ij}-\mu_{ij}=0,\forall i=1,...,m,\forall j=1,...,d
\end{flalign}
the dual problem of (\ref{eq4}) is obtained as follows,
\begin{flalign}
    \mathop{max}\limits_{\eta,\theta}\ &-\frac{1}{2}\sum\limits_{i=1}^{m}\sum\limits_{k=1}^{m}\sum\limits_{j=1}^{d}\eta_{ij}\eta_{kj}y_{ij}y_{kj}\Phi(x_{i},j)\cdot\Phi(x_{k},j)\nonumber \\
    &+\sum\limits_{i=1}^{m}\sum\limits_{k=1}^{m}\sum\limits_{j=1}^{d}\eta_{ij}\theta_{kj}y_{ij}(1+y_{kj})\Phi(x_{i},j)\cdot\Phi(x_{k},j)\nonumber \\
    &+\frac{1}{2}\sum\limits_{i=1}^{m}\sum\limits_{k=1}^{m}\sum\limits_{j=1}^{d}\theta_{ij}\theta_{kj}(1+y_{ij})(1+y_{kj})\Phi(x_{i},j)\cdot\Phi(x_{k},j) \nonumber \\
    &+\sum\limits_{i=1}^{m}\sum\limits_{j=1}^{d}\eta_{ij} \nonumber \\
    s.t. &\sum\limits_{i=1}^{m}\eta_{ij}y_{ij}=0,\forall j=1,...,d \nonumber \\
    &\sum\limits_{i=1}^{m}\theta_{ij}(1+y_{ij})=0,\forall j=1,...,d \nonumber \\
    &0\leq \eta_{ij} \leq C_{1},for\ i=1,...,m\ and\ j =1,...,d.
\end{flalign}
If the solution to the above problem is $(\eta_{11},...,\eta_{ij},...,\eta_{md},\theta_{11},...,\theta_{ij},...,\theta_{md})$ where is $\eta_{ij}\geq 0,y_{kj}=1\ and\ \theta_{kj}\neq -C_{2} $ ,the solution of $(W,b)$ can be calculated as following:
\begin{flalign}
    W=&\sum\limits_{i=1}^{m}\sum\limits_{j=1}^{d}\eta_{ij}y_{ij}\Phi(x_{i},j) + \sum\limits_{i=1}^{m}\sum\limits_{j=1}^{d}\theta_{ij}(1+y_{ij})\Phi(x_{i},j) \\
    b_{j1}=&y_{ij}-W\cdot \Phi(x_{i},j),\forall j=1,...,d  \\
    b_{j2}=&-(1+y_{ij})(W\cdot \Phi(x_{i},j)),\forall j=1,...,d 
\end{flalign}
\subsection{ Non linear parallel multi label support vector machine}
One of the characteristics of support vector machines which is widely used is that they can be used to estimate nonlinear functions.
We can clearly generalize the linear PMLSVM method outlined above to the non-linear case using kernels as is done for SVM.
We define nonlinear feature map
\begin{flalign}
    \emptyset\ : \Phi(x_{i},j) \rightarrow H
\end{flalign}
where $H$ is a separable Hilbert space.The kernel associated to $\emptyset$ is
\begin{flalign}
    K(\Phi(x_{i},j),\Phi(x_{k},s))=<\emptyset(\Phi(x_{i},j)),\emptyset(\Phi(x_{k},s))>
\end{flalign}
where $<.,.>$ is the inner product in $H$.The kernel method is introduced into the dual problem to obtain
\begin{flalign}
    \mathop{max}\limits_{\eta,\theta}\ &-\frac{1}{2}\sum\limits_{i=1}^{m}\sum\limits_{k=1}^{m}\sum\limits_{j=1}^{d}\eta_{ij}\eta_{kj}y_{ij}y_{kj}K(\Phi(x_{i},j)\cdot\Phi(x_{k},j))\nonumber \\
    &+\sum\limits_{i=1}^{m}\sum\limits_{k=1}^{m}\sum\limits_{j=1}^{d}\eta_{ij}\theta_{kj}y_{ij}(1+y_{kj})K(\Phi(x_{i},j)\cdot\Phi(x_{k},j))\nonumber \\
    &+\frac{1}{2}\sum\limits_{i=1}^{m}\sum\limits_{k=1}^{m}\sum\limits_{j=1}^{d}\theta_{ij}\theta_{kj}(1+y_{ij})(1+y_{kj})K(\Phi(x_{i},j)\cdot\Phi(x_{k},j)) \nonumber \\
    &+\sum\limits_{i=1}^{m}\sum\limits_{j=1}^{d}\eta_{ij} \nonumber \\
    s.t. &\sum\limits_{i=1}^{m}\eta_{ij}y_{ij}=0,\forall j=1,...,d \nonumber \\
    &\sum\limits_{i=1}^{m}\theta_{ij}(1+y_{ij})=0,\forall j=1,...,d \nonumber \\
    &0\leq \eta_{ij} \leq C_{1},for\ i=1,...,m\ and\ j =1,...,d
\end{flalign}
\section{Experiments}
\subsection{Experimental setup}
To evaluate the performance of MLPSVM,in this section we investigate its performance on real-world datasets.
As a comparison, we compare MLPSVM with other five multi-label classifiers.
Including MLKNN\cite{MLKNN},MLARAM\cite{MLARAM},BR\_SVM,CC\_SVM\cite{read2009classifier},MLTSVM\cite{MLTSVM}.
In order to evaluate the performance of the algorithm, we select the following four metrics
including Hmloss(Hamming loss),Oerr(One-error),Pre(Precision),Rec(Recall).
Let's introduce these four metrics.
\begin{itemize}
\item Hamming loss: Evaluates how many times an instance-label pair is misclassified between the 
pre dicted label set $h(x)$ and the ground-truth label set y
\begin{flalign}
    Hmloss(h)=\frac{1}{p}\sum\limits_{i=1}^{p}\frac{1}{k}|h(x_{i})\Delta y_{i}|\in [0,1],
\end{flalign}
where $\delta$ stands for the symmetric difference of two sets.\\
\item One-error:Evaluate the number of times top-ranked label is not in the sample's real label set
the smaller the value, the better the performance.
\begin{flalign}
    one-error(h)=\frac{1}{p}\sum\limits_{i=1}^{p}h(x_{i})\in [0,1],h(x_{i})=\bigg \{_{1\ otherwise}^{0\ if\ arg maxf_{y}(x_{i})\in y_{i}}
\end{flalign}
\item Precision,Recall:
\begin{flalign}
    Precision(h)=\frac{1}{p}\sum\limits_{i=1}^{p}\frac{|Y_{i}\cap h(x_{i})|}{|h(x_{i})|}\\
    Recall(h)=\frac{1}{p}\sum\limits_{i=1}^{p}\frac{|Y_{i}\cap h(x_{i})|}{|Y_{i}|}
\end{flalign}
\end{itemize}
All the experiments are done on personal computers with an Intel Core-i5 7400 processor(3.00GHz) and 8 GB random access memory(RAM).
All comparison algorithms are from python's third repository Scikit-Multilearn\cite{2017arXiv170201460S}.
We selected two continuous data sets from MULAN\cite{Tsoumakas2009Mining} multi-label learning librarie.As shown in Table\ref{dataset}.
These datasets represent a wide range of domains(audio,image,biology and music).
\begin{table}[htb]
    \centering
    \caption{Real world dataset introduction from MULAN multi-label learning open source library}
    \label{dataset}
    \begin{center}
    \begin{tabular}{ccccc}
        \toprule
        Dataset  & Domain & Instances & Features & Labels \\ \midrule
        CAL500   & music  & 502       & 68       & 174    \\
        Scene    & image  & 2407      & 294      & 6      \\
        Birds    & audio  & 645       & 258      & 21-1   \\
        Yeast    & biology& 2417      & 103      & 14     \\
        Emotions & music  & 593       & 72       & 6      \\ \bottomrule
        \end{tabular}
    \end{center}
    \end{table}
\subsection{Results on real world datasets}
In this section, linear MLPSVM is compared with other multi-label learning algorithms.
In contrast, other algorithm that use SVM as that base classifier will also use linear SVM.
Table \ref{Music},\ref{Emotions},\ref{Birds},\ref{Scene}\ and\ \ref{Yeast}\ show the experimental results.
The experimental results show that MLPSVM performs well on Emotions and Yeast data sets,which may be the cross distribution of data.
The performance of MLPSVM on one\_error evaluation index is obviously better than other algorithms.
\begin{table}[htb]
    \caption{Results of Multi-label Classification Algorithm on CAL500 Data}
    \label{Music}
    \begin{center}
    \begin{tabular}{ccccc} \toprule
    ~ & \multicolumn{4}{c}{CAL500} \\ \cmidrule{2-5}
    Algithms &  Hmloss               &             Oerr     &              Pre        &             Arc \\ \midrule
    MLPSVM &    0.141$\pm$0.003      & \bf{0.002$\pm$0.018} &0.572  $\pm$0.018        &  0.223$\pm$0.010 \\
    MlARAM &    0.181$\pm$0.019      &0.388$\pm$0.124       &0.376  $\pm$0.134        &	0.203$\pm$0.069\\
    MLkNN  &    0.145$\pm$0.005      &0.140$\pm$0.060       &0.530  $\pm$0.029        &  \bf{0.264$\pm$0.020}\\
    MLTSVM &	0.196$\pm$0.006      &\_                    &0.103  $\pm$0.026        &	0.040$\pm$0.010 \\
    BR\_SVM&	\bf{0.137$\pm$0.004} &0.115 $\pm$0.051      &\bf{0.618  $\pm$0.021}   &	0.226$\pm$0.009 \\
    CC\_SVM&	\bf{0.137$\pm$0.004} &0.145 $\pm$0.034      &0.613  $\pm$0.013        & 	0.223$\pm$0.006\\\bottomrule 
    \end{tabular}
    \end{center}
\end{table}

\begin{table}
    \caption{Results of Multi-label Classification Algorithm on Emotions Data}
    \label{Emotions}
    \begin{center}
    \begin{tabular}{ccccc} \toprule
    ~ & \multicolumn{4}{c}{Emotions} \\ \cmidrule{2-5}
    Algithms &              Hmloss     &               Oerr        &               Pre         &         Arc \\ \midrule
    MLPSVM	&   \bf{0.213} $\pm$0.015  &	\bf{0.062 $\pm$	0.024} &	\bf{0.698 $\pm$	0.039} &	\bf{0.576 $\pm$	0.034} \\
    MLARAM	&	0.363 $\pm$	0.032      &	0.531 $\pm$	0.078      &	0.422 $\pm$	0.051      &	0.396 $\pm$	0.056 \\
    MLkNN	&	0.265 $\pm$	0.021      &	0.371 $\pm$	0.043      &	0.604 $\pm$	0.054      &	0.440 $\pm$	0.049 \\
    MLTSVM	&	0.243 $\pm$	0.011      &	\_                     &	0.645 $\pm$	0.029      &	0.495 $\pm$	0.030 \\
    BR\_SVM	&	0.244 $\pm$	0.026      &	0.344 $\pm$	0.042      &	0.664 $\pm$	0.062      &	0.438 $\pm$	0.049 \\
    CC\_SVM	&	0.240 $\pm$	0.035      &	0.355 $\pm$	0.060      &	0.673 $\pm$0.071       &	0.452 $\pm$	0.074 \\\bottomrule 
    \end{tabular}
\end{center}
\end{table}

\begin{table}
    \caption{Results of Multi-label Classification Algorithm on Birds Data}
    \label{Birds}
    \begin{center}
    \begin{tabular}{ccccc} \toprule
    ~ & \multicolumn{4}{c}{Birds} \\ \cmidrule{2-5} 
    Algithms &              Hmloss     &               Oerr        &               Pre         & Arc \\ \midrule
    MLPSVM	 &	0.062 $\pm$	0.007      &	\bf{0.003 $\pm$	0.006} &	0.396 $\pm$	0.110      &	0.256 $\pm$	0.066 \\
    MLARAM   &	0.079 $\pm$	0.005      &	0.523 $\pm$	0.046      &	0.462 $\pm$	0.061      &0.352  $\pm$	0.032 \\
    MLkNN    &	\bf{0.053 $\pm$	0.006} &	0.014 $\pm$	0.013      &	\bf{0.876 $\pm$	0.067} &	0.341 $\pm$	0.025 \\
    MLTSVM   &  0.194 $\pm$	0.017      &	\_                     &	0.158 $\pm$	0.017      &	\bf{0.367 $\pm$	0.032} \\
   BR\_SVM   &  0.056 $\pm$	0.007      &	0.101 $\pm$	0.039      &	0.818 $\pm$	0.068      &	0.320 $\pm$	0.018 \\
   CC\_SVM	 &	0.056 $\pm$	0.007      &	0.104 $\pm$	0.028      &	0.812 $\pm$	0.061      &	0.321 $\pm$	0.028 \\\bottomrule 
    \end{tabular}
\end{center}
\end{table}

\begin{table}
    \caption{Results of Multi-label Classification Algorithm on Scene Data}
    \label{Scene}
    \begin{center}
    \begin{tabular}{ccccc} \toprule
    ~ & \multicolumn{4}{c}{Scene} \\ \cmidrule{2-5}
    Algithms&               Hmloss     &               Oerr        &               Pre         & Arc \\ \midrule
    MLPSVM	&	0.173 $\pm$	0.010      &	\bf{0.035 $\pm$	0.017} &	0.532 $\pm$	0.046      &	0.420 $\pm$	0.039 \\
    MLARAM	&   0.115 $\pm$	0.011      &	0.227 $\pm$	0.040      &	0.643 $\pm$	0.025      &	\bf{0.820 $\pm$	0.036} \\
    MLkNN	&   \bf{0.092 $\pm$	0.006} &	0.208 $\pm$	0.031      &	0.782 $\pm$	0.024      &	0.684 $\pm$	0.036 \\
    MLTSVM  &   0.164 $\pm$	0.009      &	\_                     &	0.569 $\pm$	0.037      &	0.392 $\pm$	0.031 \\
    BR\_SVM &	0.142 $\pm$	0.005      &	0.531 $\pm$	0.039      &	\bf{0.953 $\pm$	0.034} &	0.226 $\pm$	0.036 \\
    CC\_SVM	&	0.137 $\pm$	0.007      &	0.447 $\pm$	0.047      &	0.942 $\pm$	0.033      &	0.262 $\pm$	0.034 \\\bottomrule 
    \end{tabular}
    \end{center}
\end{table}

\begin{table}
    \caption{Results of Multi-label Classification Algorithm on Yeast Data}\
    \label{Yeast}
    \begin{center}
    \begin{tabular}{ccccc} \toprule
    ~       & \multicolumn{4}{c}{Yeast} \\ \cmidrule{2-5}
    Algithms&               Hmloss     &Oerr                        &               Pre         &   Arc \\ \midrule
    MLPSVM  &   \bf{0.187 $\pm$	0.010} &\bf{0.062 $\pm$	0.019}      &	0.712 $\pm$	0.026       &	0.570 $\pm$	0.020 \\
    MLARAM	&	0.217 $\pm$	0.010      &0.300 $\pm$	0.023           &	0.653 $\pm$	0.023       &	\bf{0.606 $\pm$	0.019} \\
    MLkNN	&	0.199 $\pm$	0.010      &0.248 $\pm$	0.022           &	0.706 $\pm$	0.022       &	0.589 $\pm$	0.015 \\
    MLTSVM	&	0.311 $\pm$	0.004      &	\_                      &	0.408 $\pm$	0.035       &	0.060 $\pm$	0.012 \\
    BR\_SVM	&   0.225 $\pm$	0.008      &0.389 $\pm$	0.032           &	\bf{0.755 $\pm$	0.025}  &	0.378 $\pm$	0.013 \\
    CC\_SVM	&   0.236 $\pm$	0.007      &0.380 $\pm$	0.055           &	0.696 $\pm$	0.031       &	0.394 $\pm$	0.022 \\\bottomrule 
    \end{tabular}
    \end{center}
\end{table}
\section{Discussion}
MLPSVM provides a new idea for processing multi-label data.
Applying MLPSVM to real data sets, the results show that MLPSVM has better effect on Emotions and Yeast data sets.
At the same time, the results show that MLPSVM performs poorly on Birds and Scene data sets.
Experiments show that MLPSVM can process specific data well.After analysis MLPSVM has the following two areas to be improved:
\begin{itemize}
\item MLPSVM uses two hyperplanes, which will degrade the performance when the distribution of training samples is not true.
\item A fast solution method for MLPSVM has not been found, which will limit MLSVM to solve large-scale problems.
\end{itemize}
\bibliographystyle{unsrt}
\bibliography{pmlsvm}
\end{document}